\documentclass[letterpaper]{article} 
\usepackage{aaai2026}  
\usepackage{times}  
\usepackage{helvet}  
\usepackage{courier}  
\usepackage[hyphens]{url}  
\usepackage{graphicx} 
\usepackage{natbib}  
\usepackage{caption} 
\usepackage{subcaption}
\usepackage{amsmath}
\usepackage{enumitem}
\usepackage{newfloat}
\usepackage{listings}
\usepackage{multirow} 
\DeclareCaptionStyle{ruled}{labelfont=normalfont,labelsep=colon,strut=off} 
\usepackage{float}
\usepackage[table]{xcolor}
\floatstyle{ruled}                 
\newfloat{myfloat}{tb}{myf}        
\floatname{myfloat}{Listing}       
\usepackage{booktabs} 
\usepackage{amssymb}
{
}
\title{\textsc{ReMisVFU}: Vertical Federated Unlearning via Representation Misdirection for Intermediate Output Feature}
\author{
    Wenhan Wu\textsuperscript{\rm 1,\equalcontrib},
    Zhili He\textsuperscript{\rm 1,2,\equalcontrib},
    Huanghuang Liang\textsuperscript{\rm 1},
    Yili Gong\textsuperscript{\rm 1},
    Jiawei Jiang\textsuperscript{\rm 1},
    Chuang Hu\textsuperscript{\rm 3,4,\corrds},
    Dazhao Cheng\textsuperscript{\rm 1,\corrds}
}
\affiliations{
    \textsuperscript{\rm 1}School of Computer Science, Wuhan University, Wuhan, China\\
    \textsuperscript{\rm 2}Southwest Aluminium (GROUP) Co., Ltd., Chongqing, China\\
    \textsuperscript{\rm 3}State Key Laboratory of Internet of Things for Smart City, University of Macau, Macau SAR\\
    \textsuperscript{\rm 4}Key Laboratory of Transportation Internet of Things, Wuhan University of Technology, Wuhan, China

    \{wenhanwu, 2022182110069, hhliang, yiligong, jiawei.jiang, dcheng\}@whu.edu.cn, chuanghu@um.edu.mo
}

\usepackage{bibentry}
\usepackage{subcaption}
\definecolor{lightgray}{gray}{0.75}          
\newcommand{\best}[1]{\cellcolor{lightgray}{#1}}
\usepackage[ruled,vlined,linesnumbered]{algorithm2e}
\SetKwComment{tcp}{// }{}       
\SetKwInput{KwIn}{Input}        
\SetKwInput{KwOut}{Output}      
\SetAlgoLined                  
\begin{document}
\maketitle

\begin{abstract}
Data-protection regulations such as the GDPR grant every participant in a federated system a right to be forgotten. Federated unlearning has therefore emerged as a research frontier, aiming to remove a specific party's contribution from the learned model while preserving the utility of the remaining parties. However, most unlearning techniques focus on Horizontal Federated Learning (HFL), where data are partitioned by samples. In contrast, Vertical Federated Learning (VFL) allows organizations that possess complementary feature spaces to train a joint model without sharing raw data. The resulting feature-partitioned architecture renders HFL-oriented unlearning methods ineffective. In this paper, we propose \textsc{ReMisVFU}, a plug-and-play representation misdirection framework that enables fast, client-level unlearning in splitVFL systems. When a deletion request arrives, the forgetting party collapses its encoder output to a randomly sampled anchor on the unit sphere, severing the statistical link between its features and the global model. To maintain utility for the remaining parties, the server jointly optimizes a retention loss and a forgetting loss, aligning their gradients via orthogonal projection to eliminate destructive interference. Evaluations on public benchmarks show that \textsc{ReMisVFU} suppresses back-door attack success to the natural class-prior level and sacrifices only about 2.5\% points of clean accuracy, outperforming state-of-the-art baselines.
\end{abstract}

\section{Introduction}
Federated Learning (FL)~\cite{mcmahan2017communication} allows multiple clients to train a shared model while keeping raw data local, exchanging only encrypted gradients or intermediate representations. Existing mainstream paradigm of FL is horizontal federated learning (HFL)~\cite{yang2020horizontal}, which assumes all parties own data drawn from the same feature space, yet different samples. However, this design struggles when collaborating organizations possess heterogeneous feature spaces. To address the limitation, Vertical Federated Learning (VFL)~\cite{cheng2021secureboost,liu2024vertical} was proposed. The VFL parties refer to the same set of users, but each controls a disjoint subset of features. Given a typical cross-industry scenario, an e-commerce platform that holds customers' purchase histories and a bank that records their income and credit scores can train a credit-scoring model without revealing their respective raw data via VFL, thereby reducing privacy-leakage risk.

Recently, with the enforcement of the EU GDPR~\cite{voigt2017eu} and the California CCPA~\cite{bonta2022california}, the \textit{right to be forgotten}, i.e., the ability to erase specific data from an ML system, has become a key legal requirement. This mandate has spurred research on \textit{Federated Unlearning (FU)}~\cite{liu2020federated,varshney2025unlearning,liu2024survey,ijcai2025p733}, the federated counterpart of machine unlearning. FU seeks to adjust an already-trained model so that, once a user's erasure request is honored, the resulting model behaves as if that data had never been seen. Although retraining the model from scratch would technically meet the requirement, it is prohibitively time and resource-intensive. Consequently, HFL unlearning (HFU) research has focused on more efficient alternatives, such as reusing stored gradients~\cite{liu2021federaser,liu2022right}, applying knowledge distillation~\cite{zhu2023heterogeneous,wu2025mimir,xie2024adaptive} to remove or preserve information selectively, and performing gradient-ascent updates that directly counteract the departing client's contribution~\cite{pan2025federated,hua2024federated,li2023subspace}. However, nearly all existing techniques targeting HFL cannot be transferred to VFL without major redesign because the two paradigms differ fundamentally in both training and inference. State-of-the-art VFL systems, such as splitVFL~\cite{gupta2018distributed}, partition a model into multiple local (bottom) models hosted by passive parties (feature holders) and a global (top) model hosted by the active party (label holder). During training, each bottom model forwards intermediate output features to the top model and then completes forward and back-propagation; inference likewise requires all parties to collaborate. Therefore, we identify two unique challenges in VFL unlearning:
\begin{itemize}
    \item \textbf{C1. Data is vertically partitioned}: Unlike HFU, which directly removes entire labeled samples, VFU targets only the feature slice contributed by the departing party, whose influence on prediction is more concealed within the model and more difficult to eliminate completely.
    \item \textbf{C2. Model partitioning and collaborative inference}: A departing party's features affect both its local bottom model and the active party's top model, so unlearning must revise both components and reorganize the inference workflow to fit the updated VFL architecture.
\end{itemize}

To tackle these challenges, we propose a novel VFU framework named \textsc{ReMisVFU} (\underline{\textbf{Re}}presentation \underline{\textbf{Mis}}direction for \underline{\textbf{V}}ertical \underline{\textbf{F}}ederated \underline{\textbf{U}}nlearning). Instead of retraining the model from scratch, \textsc{ReMisVFU} performs representation-level misdirection: the forgetting party replaces its intermediate features with anchor vectors sampled from a uniform sphere, which retains the information flow from the departing party while preserving the original VFL pipeline. To ensure utility preservation, we jointly optimize forgetting and retention objectives using a gradient projection strategy that resolves conflicts between their respective gradients. Our key contributions are summarized as follows:
\begin{itemize}
\item We propose a representation misdirection mechanism that effectively erases the statistical contribution of the forgetting party by collapsing its output features.

\item We design a coordination strategy that ensures retention gradients are orthogonal to forgetting gradients, preventing mutual interference and enabling stable optimization.

\item We develop a plug-and-play unlearning pipeline fully compatible with existing splitVFL systems, requiring no changes to communication patterns or party workflows.

\item Comprehensive evaluations demonstrate that \textsc{ReMisVFU} achieves strong performance both on forgetting and retention objectives.
\end{itemize}

\section{Preliminary and Related Works}

\subsection{Vertical Federated Learning}
VFL departs fundamentally from the HFL paradigm.  
In HFL every client retains a local subset of samples that share an identical feature space.  
In VFL, by contrast, each party owns a subset of features describing an overlapping subset of the same individuals.  
We first formalize the VFL setting adopted in the paper.
We denote a dataset with $N$ samples:
\begin{equation}
    \mathcal{D}=\bigl\{(x_i,y_i)\bigr\}_{i=1}^{N},
\qquad
x_i\in\mathbb{R}^{1\times d},\;
y_i\in\mathcal{Y},
\end{equation}
VFL aims to collaboratively train a global model $M$ while preserving each party's data privacy and model security.  
Formally, we solve the above optimization problem:
\begin{equation}
\min_{\Omega}\;
\mathcal{L}(\Omega;\mathcal{D})
=
\frac{1}{N}\sum_{i=1}^{N}
f\bigl((x_i,y_i);\Omega\bigr)
+
\lambda\sum_{k=1}^{K}\gamma(\Omega_k),
\label{eq:global_obj}
\end{equation}
where $\Omega=\{\Omega_1,\dots,\Omega_{K}\}$ collects all trainable parameters,  
$f(\cdot)$ is the task loss (e.g.\ cross‐entropy for classification task),  
$\gamma(\cdot)$ is a regulariser, and hyperparameter $\lambda>0$ balances the two terms. Each sample's feature vector is column-wise split among $K$ parties $P_1,\dots,P_K$:
\begin{equation}
x_i
=\bigl[\,
x_{i,1}\,\|\,x_{i,2}\,\|\dots\|\,x_{i,K}
\bigr],
\end{equation}
where $x_{i,k}\in\mathbb{R}^{1\times d_k}$ and $\sum_{k=1}^{K}d_k=d$. $d_k$ is the
feature dimension of party $k \in [1,K]$. One designated \emph{active} party—without loss of generality $P_K$—additionally retains the label, i.e.\ $y_i=y_{i,K}$.  
Thus the local dataset for the passive party $P_k, k \in [1,K-1]$ is:
\begin{equation}
\mathcal{D}_k=\{x_{i,k}\}_{i=1}^{N},\quad k\in[1,K\!-\!1],
\end{equation}
and the local dataset for the active party $P_K$ is:
\begin{equation}
\mathcal{D}_K=\bigl\{(x_{i,K},y_{i,K})\bigr\}_{i=1}^{N}.
\end{equation}
Then we can decompose the global model into bottom models ${\rm ML}_k(\cdot;\omega_k)$ hosted on $P_k$ ($k=1,\dots,K$) and a top module ${\rm MG}_K(\cdot;\varphi_K)$ that resides only on the active party.  
With $\Omega_k=\omega_k$ for $k<K$ and $\Omega_K=(\omega_K,\varphi_K)$, the per-sample loss can be rewritten as:
\begin{equation}
\small
\begin{aligned}
    &f\bigl((x_i,y_i);\Omega\bigr)
\\ = &
\ell\!\Bigl(
{\rm MG}_K\!\bigl(
{\rm ML}_1(x_{i,1};\Omega_1),\dots,{\rm ML}_K(x_{i,K};\Omega_K)
;\varphi_K\bigr),
\,y_{i,K}
\Bigr),
\end{aligned}
\label{eq:sample_loss}
\end{equation}
where $\ell(\cdot,\cdot)$ denotes the task-specific loss. Eq.~(\ref{eq:sample_loss}) assumes that the active party $P_K$ owns a subset of features $x_{i,K}$ in addition to the labels $y_{i,K}$. However, in many real-world deployments, the entity holding the labels is not a data provider but merely an annotator (e.g., a hospital supplying diagnoses, or a third-party rating agency).  
In this label-only setting, we simply let $d_K=0$ so that  
$x_{i,K}=\varnothing$ and the active party contributes no local encoder ${\rm ML}_K$.  
Consequently, $f\bigl((x_i,y_i);\Omega\bigr)$ in Eq.~(\ref{eq:sample_loss}) reduces to:
\begin{equation}
\small
\ell\!\Bigl({\rm ML}_1(x_{i,1};\Omega_1),\dots,
{\rm ML}_{K-1}(x_{i,K-1};\Omega_{K-1})
;\varphi_K\bigr),
\,y_{i,K}
\Bigr),
\label{eq:sample_loss2}
\end{equation}
where ${\rm MG}_K$ aggregates only the encrypted representations received from the passive parties.  
A VFL framework accommodates both feature-rich and label-only active parties. The training round proceeds as follows:  
First, the parties privately align entity IDs \cite{zhao2024deep}.  
Each party then performs a local forward pass and obtains its immediate hidden representation $h_{i,k}={\rm ML}_k(x_{i,k};\Omega_k)$.  
Passive parties $P_k$ encrypt (or otherwise protect) $\{h_{i,k}\}$ and transfer them to the active party $P_K$.  
The active party decrypts and aggregates the representations through ${\rm MG}_K$, evaluates the loss using Eq.~(\ref{eq:sample_loss}), computes gradients, and updates $\varphi_K$ together with all $\Omega_k$.  
The refreshed $\Omega_k$ ($k<K$) are finally returned to their respective owners.  
Repeating these VFL rounds until convergence achieves joint training while ensuring that raw features and labels remain local.
For the model architecture, 
When deep models are employed, the global module ${\rm MG}_K$ usually comprises trainable network layers, in a manner reminiscent of vertical SplitNN~\cite{vepakomma2018split,ceballos2020splitnn}, where the entire model is partitioned layer-wise across different parties. This configuration is commonly named splitVFL~\cite{gupta2018distributed}, one of the most popular VFL architectures. In this paper, we focus on splitVFL as the primary object of study.

\subsection{Vertical Federated Unlearning}
For the vertical federated unlearning (VFU) problem, the goal is to remove the influence of designated feature owners from an already trained model while preserving the utility for all remaining parties. 
\cite{deng2023vertical} were the first to articulate the concept, though their method is restricted to logistic-regression models. Similarly, \cite{li2024securecut} designed a VFU framework tailored to federated gradient-boosted decision trees. For the deep learning model, \cite{wang2024efficient} enables rapid retraining in VFL by retaining checkpoints of lower-layer models, but at the cost of significant storage overhead. \cite{varshney2025unlearning} adopts a knowledge-distillation approach that focuses on changing the model architecture rather than truly removing the memorized information. Building on these studies, we formulate the client-level VFU process as follows:

Let $P_f$ be the party that requests its contributions to be forgotten.   
Its local data are denoted by $\mathcal{D}_f=\{x_{i,f}\}_{i=1}^{N}$, where $x_{i,f}\in\mathbb{R}^{1\times d_f}$ and $\sum_{k\neq f}d_k+d_f=d$.  
The remaining parties $\mathcal{P}_r=\{P_k\}_{k\neq f}$ with data $\mathcal{D}_r=\bigl\{x_{i,k}\bigr\}_{k\neq f,i=1}^{N}$ constitute the retained side. It is worth emphasizing that, irrespective of whether the forgetting party $P_f$ acts as the active party, our study deliberately excludes the erasure of label data, because removing the labels would strip the model of its supervisory signal and thus preclude further execution of the classification task.

\noindent\textbf{Baseline (retrained) model.}
Ideally, the system would discard all parameters related to $\mathcal{D}_f$ and retrain splitVFL from scratch using only the retained parties:
\begin{equation}
\Omega^\star \;=\;
\mathcal{VFL}\bigl(\mathcal{D}\setminus\mathcal{D}_f\bigr),
\label{eq:vfl_retrain}
\end{equation}
where $\mathcal{VFL}$ denotes the standard training pipeline described in Figure~\ref{fig:vfl_framework}.  
Model $\Omega^\star$ serves as the \textit{gold standard} for evaluating unlearning quality~\cite{wang2022federated,meerza2024confuse}.

\noindent\textbf{Unlearning process.}
In practice, it is often impractical to perform full retraining because of the high resource cost. We therefore introduce a VFU operator as follows:
\begin{equation}
\mathcal{VU}: \;
\mathcal{VFL}(\mathcal{D}) \;\otimes\; \mathcal{D}_r \;\otimes\; \mathcal{D}_f
\;\longrightarrow\;
\widetilde{\Omega},
\end{equation}
which takes the original splitVFL model $\mathcal{VFL}(\mathcal{D})$, the complete retained data $\mathcal{D}_r$ (only locally available at each party), and the forgetting partition~$\mathcal{D}_f$ as input, and outputs an unlearned parameter set~$\widetilde{\Omega}$.

\noindent\textbf{Unlearning objective.}
VFU is successful when the distribution of predictions made by $\widetilde{\Omega}$ is approximately indistinguishable from that of the baseline model in Eq.~(\ref{eq:vfl_retrain}):
\begin{equation}
\Phi\!\bigl[\,
\mathcal{VFL}(\mathcal{D}\setminus\mathcal{D}_f);\Omega^*
\bigr]
\;\approx\;
\Phi\!\bigl[\,
\mathcal{VU}\!\bigl(\mathcal{VFL}(\mathcal{D}),\,\mathcal{D},\,\mathcal{D}_f\bigr); \widetilde{\Omega}
\bigr],
\label{eq:vfu_goal}
\end{equation}
where $\Phi[\cdot]$ denotes the predictive distribution of a model. Kullback–Leibler divergence can quantify the discrepancy between the two distributions \cite{kullback1951information}:
\begin{equation}
\mathrm{KL}\!\bigl(
\Phi^\star
\;\|\;
\widetilde{\Phi}
\bigr)
\;=\;
\mathbb{E}_{x\sim\Phi^\star}
\!\left[
\log
\Phi^\star(x) / \widetilde{\Phi}(x)
\right],
\label{eq:kl}
\end{equation}
where $\Phi^\star=\Phi\bigl[\mathcal{VFL}(\mathcal{D}\setminus\mathcal{D}_f)\bigr]$ and $\widetilde{\Phi}=\Phi\bigl[\widetilde{\Omega}\bigr]$.  
An effective unlearning algorithm seeks to minimize Eq.(\ref{eq:kl}) with minimal resource and time overhead.

\section{Methodology}
\newlength{\oldtextfloatsep}
\setlength{\oldtextfloatsep}{\textfloatsep} 

\subsection{System Overview}
\textsc{ReMisVFU} is a plug-and-play unlearning module for splitVFL. When the forgetting party $P_f$ submits an erasure request, we apply representation misdirection to the intermediate output features produced by its bottom model/encoder ${\rm ML}_f$, which means $P_f$ substitutes its informative representation $h_{i,f}$ with a target anchor derived from a random vector. This procedure removes the influence of $P_f$'s representations on the global model while requiring only a few additional training epochs. The active party $P_K$ harmonizes the resulting gradients between the retained gradients and forgotten gradients, and all parameters $\Omega$ are updated collaboratively, achieving fast erasure with virtually no significant drop in overall accuracy. 

We study a classification-oriented deep VFL architecture that concatenates the intermediate feature representations produced by multiple parties and feeds them into a top aggregator $\mathrm{MG}_k$ to generate the final prediction. Our objective is twofold: (i) suppress the model's capacity to extract information originating from the forgotten party $P_f$, and (ii) simultaneously preserve its ability to exploit the information contributed by the retained parties $P_r\;(r\neq k)$. 

For the unlearning procedure, we depart from conventional class-level unlearning that targets accuracy metrics directly~\cite{chundawat2023zero,chen2023boundary}. Instead, we pursue a more generalizable approach to erase the entire knowledge distribution contributed by $P_f$. Each VFU round reuses the exact communication pattern of standard VFL, leaving the workflow of passive parties $P_{k\neq f}$ completely unaffected; hence, the method can be integrated into existing VFL frameworks without any protocol modification.

\begin{figure}
    \centering
    \includegraphics[width=0.8\linewidth]{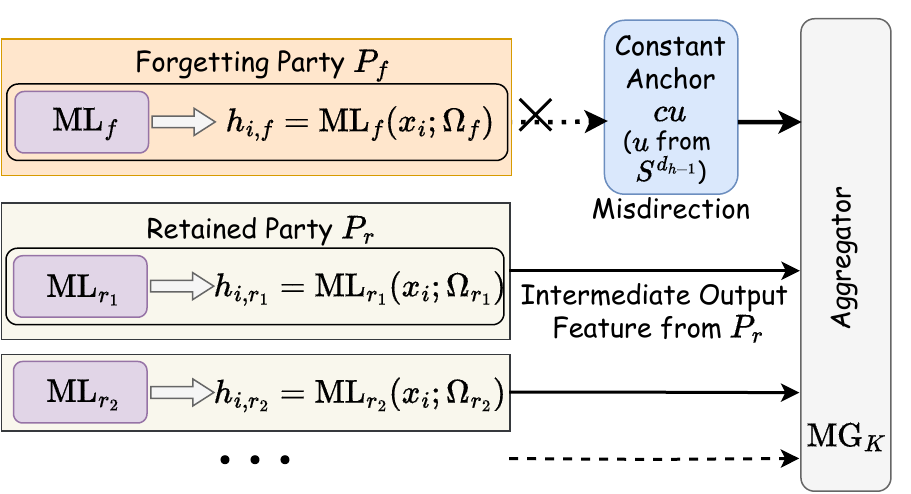}
    \caption{Representation Misdirection Unlearning in VFL.}
    \label{fig:rmu}
\end{figure}

\subsection{Representation Misdirection (Forgetting) Loss \(\mathcal{L}_{f}\)}
Traditionally, deep neural networks are trained by minimizing a loss that depends solely on their final outputs. Mechanistic interpretability proposes editing a model by intervening on individual neurons \cite{mazzia2024survey,mitchell2021fast}. Departing from both perspectives, we adopt the view that a model's internal representations already encode structured world knowledge and can therefore be deliberately manipulated to steer the model's behavior \cite{wang2023machine,li2024wmdp,wu2025defending}. In VFL, the intermediate features emitted by each party naturally serve as such representations and can be harnessed for unlearning. Building on this insight, we devise a representation-misdirection forgetting loss that injects controlled perturbations into the features contributed by the forget party $P_f$ based on random alignment. Intuitively, this loss scrambles the intermediate outputs of $P_f$ so thoroughly that the top aggregator $\mathrm{MG}_K$ can no longer recover any useful information from that party, which achieves the unlearning objectives.

More specifically, as shown in Figure \ref{fig:rmu}, to compute the forgetting loss, we have access to $h_{i,f}$, i.e., the latent feature representations $\mathrm{ML}_f(x_{i,f};\Omega_f)$ produced by the forget party $P_f$'s bottom model for an input sample $x_{i,f}$. To obliterate the statistical footprint of \(\mathcal{D}_{f}\), we draw once a unit vector \(\mathbf{u}\sim\mathcal{U}(\mathbb{S}^{d_h-1})\) and force every latent vector \(h_{i,f}\) towards the constant anchor \(c\mathbf{u}\). $c$ is a hyperparameter that controls the scaling of the intermediate output features. $d_h$ denotes the dimensionality of the intermediate representation, i.e., $d_h = \dim\ (h_{i,f})$.
We define:
\begin{equation}
\mathbb{S}^{d_h-1}
=\left\{\mathbf{v}\in\mathbb{R}^{d_h}\mid \lVert\mathbf{v}\rVert_2=1\right\},
\end{equation}
the unit $(d_h-1)$-sphere in $\mathbb{R}^{d_h}$, comprising all vectors with Euclidean norm 1.
Let $\mathcal{U}(\mathbb{S}^{d_h-1})$ be the uniform distribution over this sphere.
Thus, sampling $\mathbf{u}\sim\mathcal{U}(\mathbb{S}^{d_h-1})$ means drawing a unit vector whose direction is uniformly distributed across all possible orientations while its length remains exactly 1.
Once drawn, $\mathbf{u}$ is kept fixed for the entire training process of the designated forget party $P_f$.
Given the forget dataset $\mathcal{D}_f$, the squared-\(\ell_2\) loss is as below:
\begin{equation}
\mathcal{L}_{f}
=\mathbb{E}_{x_{i,f}\sim\mathcal{D}_f}
\bigl\|h_{i,f}\;-\;c\mathbf{u}\bigr\|_2^{2},
\label{loss1}
\end{equation}
$\mathcal{L}_{f}$ quickly collapses the encoder's output manifold, guaranteeing that no membership or attribute information about \(P_f\)'s data can be recovered from the new representation.


\subsection{Utility Preservation (Retention) Loss \(\mathcal{L}_{r}\)}
The retention aim is to curb the general capability loss incurred by unlearning. While \(\mathcal{L}_{f}\) enforces forgetting, a concurrent utility loss \(\mathcal{L}_{r}\) safeguards the service quality for the remaining parties.
Rather than minimizing surrogate objectives such as KL divergence, we directly optimize the task-native loss. We reuse the original splitVFL forward path and compute the retention loss as:
\begin{equation}
\mathcal{L}_{r}
=\mathbb{E}_{(x_{i,1},\dots,x_{i,K},\,y_{i,K})\sim\mathcal{D}}
\;
\ell\!\Bigl(
\mathrm{MG}_{K}\!\bigl(\,\{h_{i,k}\}_{k=1}^{K}\bigr),
\,y_{i,K}
\Bigr),
\label{loss2}
\end{equation}
where \(\ell(\cdot)\) is either cross-entropy loss for the classification tasks or MSE loss for the regression tasks. 

\begin{figure}
    \begin{subfigure}{0.49\linewidth}
    \centering
        \includegraphics[width=\linewidth]{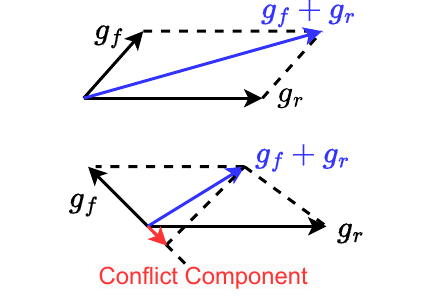}
    \caption{Update with Sum Gradients: \(g_f+g_r\)}
    \label{fig:sum_grad}
    \end{subfigure}
    \hfill
    \begin{subfigure}{0.49\linewidth}
    \centering
        \includegraphics[width=\linewidth]{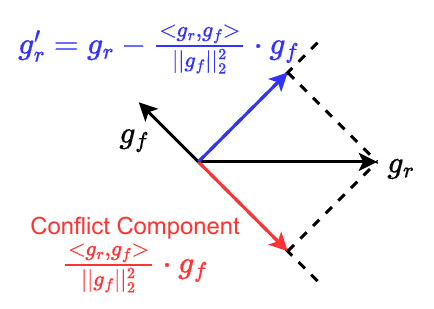}
    \caption{Update with Coordinating Gradients: \(g_f+g_r^\prime\)}
    \label{fig:coord_grad}
    \end{subfigure}
    \caption{Coordinating gradients for joint optimization. (a) Directly summing $g_r$ and $g_f$ may cause interference due to conflicting directions. (b) Projecting out the conflicting component yields $g_r^{\prime}$, which is orthogonal to $g_f$, thus avoiding interference and preserving forgetting effectiveness.}
\end{figure}

\subsection{Coordinating Gradients for Joint Optimization}
The preceding sections introduced the forgetting loss $\mathcal{L}_{f}$ and the retention loss $\mathcal{L}_{r}$. Jointly optimizing these objectives is challenging because their gradient directions frequently conflict. Strengthening the knowledge of the retained parties can inadvertently encourage the model to preserve information from the forget set $\mathcal{D}_{f}$ as shown in Figure \ref{fig:sum_grad}, thereby undermining erasure. Conversely, aggressive erasure may trigger catastrophic forgetting of transferable knowledge, ultimately diminishing downstream performance~\cite{zhao2024makes,choi2024towards}.

Motivated by insights from multi-task learning~\cite{yu2020gradient,chai2022model,huang2024learning}, we introduce a gradient–projection coordination strategy that aligns the retention and forgetting objectives in parameter space to mitigate gradient conflicts. By orthogonalizing their gradients, the optimizer follows a coherent trajectory that advances both goals efficiently. The detail is given as shown in Figure \ref{fig:coord_grad}. First, we denote the gradients of Eqs.~(\ref{loss1}) and (\ref{loss2}) for forgetting and retention, respectively:
\begin{equation}
g_{r}\;=\;\nabla_{\Omega}\,\mathcal{L}_{r},
\qquad
g_{f}\;=\;\nabla_{\Omega}\,\mathcal{L}_{f},
\end{equation}
A straightforward approach would update the model with the weighted sum as follows:
\begin{equation}
\nabla_{\Omega} \mathcal{L}_{all} =\nabla_{\Omega}\bigl(\mathcal{L}_{f}+\alpha\mathcal{L}_{r}\bigr)=g_{f}+\alpha g_{r},
\label{origrad}
\end{equation}
where $\alpha>0$ is a trade-off coefficient that scales the retention gradient, balancing preservation of utility against the strength of the forgetting signal. However, when $g_{r}$ and $g_{f}$ point in opposing directions, their components may cancel out, making forgetting sluggish and convergence unstable. To overcome the sub-optimality of naive joint optimization, we adopt a gradient-projection strategy. We measure the cosine similarity between $g_{r}$ and $g_{f}$ as below:
\begin{equation}
\cos(g_{r},g_{f})=\frac{\langle g_{r},g_{f}\rangle}{\lVert g_{r}\rVert_{2}\lVert g_{f}\rVert_{2}}.
\end{equation}
If the similarity is negative, the two objectives are locally conflicting. To ensure complete erasure and preclude interference from the retention objective with the forgetting objective, we then project $g_{r}$ onto the subspace orthogonal to $g_{f}$. By stripping $g_r$ of its conflicting component $\frac{\langle g_r, g_f\rangle}{\lVert g_f\rVert^{2}_2}\,g_f$, we obtain the adjusted gradient $\widetilde{g}_{r}$ as follows:
\begin{equation}
\widetilde{g}_{r}=
\begin{cases}
g_{r}-\dfrac{\langle g_{f},g_{r}\rangle}{\lVert g_{f}\rVert_{2}^{2}}\,g_{f},
& \langle g_{f},g_{r}\rangle<0,\\
g_{r}, & \text{otherwise},
\end{cases}
\label{cord}
\end{equation}
This orthogonality removes gradient conflicts, so updating with the projected gradient $\widetilde{g}_{r}$ incurs far less adverse impact on the forgetting objective. Conversely, if the initial cosine similarity is non-negative, the two gradient vectors are already compatible and no projection is required. The final coordinated gradient update is:
\begin{equation}
\nabla_{\Omega} \mathcal{L}_{all}^\prime =\;g_{f}+\alpha\,\widetilde{g}_{r},
\label{gradall}
\end{equation}
which retains the full forgetting signal while preserving only the non-interfering component of the retention gradient. Updates based on Eq. (\ref{gradall}) continue to drive thorough erasure of $\mathcal{D}_{f}$ without undermining the model's retention knowledge. Empirically, the projection yields faster convergence and cleaner attainment of the target distribution. The entire process is outlined in Algorithm~\ref{alg:remisvfu}.
\setlength{\algomargin}{1.5em}
\begin{algorithm}[t!]
\SetAlgoNlRelativeSize{0} 
\DontPrintSemicolon
\footnotesize
\caption{\textsc{ReMisVFU} Process}
\label{alg:remisvfu}
\KwIn{
    Pre-trained VFL model $\Omega=\{\Omega_{1},\dots,\Omega_{K}, \varphi_K\}$; Forgetting party index $f$ $(1\!\le f<K)$ with $\mathcal{D}_{f}$; Retained parties $\mathcal{P}_{r}=\{1,\dots,K\}\setminus\{f\}$ with $\mathcal{D}_{r}$; Hyper-parameters $c, \alpha$, learning rate $\eta$, epochs $T$.
}
\KwOut{Unlearned model $\widetilde{\Omega}.$}

\SetKwComment{tcp}{//}{}
\SetAlgoLined

\BlankLine
\textbf{Initialize} each party with the pretrained parameters $\Omega$ .

Draw once $\mathbf{u}\sim\mathcal{U}(\mathbb{S}^{d_h-1})$ .\hfill $\triangleright$ kept fixed

\BlankLine
\For{$t\leftarrow 1$ \KwTo $T$}{
    \ForEach{party $P_k\in\{\mathcal{P}_{r}, \mathcal{P}_{f}\}$ in parallel}{
        Sample mini-batch $\{x_{i,k}\}_{i=1}^{B}$;\\
        $h_{i,k}\leftarrow\mathrm{ML}_{k}(x_{i,k};\Omega_{k})$ .%
           \hfill$\triangleright$~~local forward
    }
    $P_K$ receives encrypted $\{h_{i,k}\}_{k=1}^{K\!-\!1}$ or $\{h_{i,k}\}_{k=1}^{K}$, decrypt, concatenate; \hfill $\triangleright$ active part aggregation\\
  
    \BlankLine
    $\hat{y}_{i}\leftarrow\mathrm{MG}_{K}\!\bigl(\{h_{i,k}\}_{k=1}^{K}\,;\varphi_{K}\bigr)$ .\hfill $\triangleright$ active part forward

    \BlankLine
    Computing forgetting loss and gradients using Eq.(\ref{loss1}): $\mathcal{L}_{f}\leftarrow\frac{1}{B}\sum_{i=1}^{B}\!\bigl\|h_{i,f}-c\mathbf{u}\bigr\|_{2}^{2},\ g_{f}\leftarrow\nabla_{\Omega}\mathcal{L}_{f}$\;  
    Computing retention loss and gradients using Eq. (\ref{loss2}): $\mathcal{L}_{r}\leftarrow\frac{1}{B}\sum_{i=1}^{B}\ell(\hat{y}_{i},y_{i,K}),\ g_{r}\leftarrow\nabla_{\Omega}\mathcal{L}_{r}$\;

    \BlankLine
    Coordinating gradients for $g_f$ and $g_r$ using Eq. (\ref{cord}):
    $\displaystyle 
    \widetilde{g}_{r}\leftarrow 
      g_{r}-\frac{\langle g_{f},g_{r}\rangle}{\lVert g_{f}\rVert_{2}^{2}}\,g_{f}\ \text{if}\ \langle g_{f},g_{r}\rangle<0$ else $g_r$\;%
      
    \BlankLine
    Updating global model $\Omega$ and broadcast to all the parties $\{\mathcal{P}_{r}, \mathcal{P}_{f}\}$:
    $\Omega\leftarrow\Omega-\eta\,(g_{f}+\alpha\,\widetilde{g}_{r})$ . 
    }

\Return $\widetilde{\Omega}\leftarrow\Omega$
\end{algorithm}

\section{Evaluations}
\subsection{Evaluation Setup}
\begin{figure}[t]
    \begin{subfigure}{0.3\linewidth}
    \centering
        \includegraphics[height=2cm]{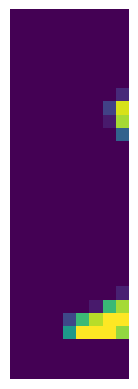}
    \caption{Left part.}
    \end{subfigure}
    \begin{subfigure}{0.3\linewidth}
    \centering
        \includegraphics[height=2cm]{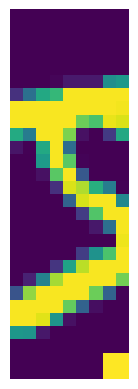}
    \caption{Centre Part.}
    \end{subfigure}
    \begin{subfigure}{0.3\linewidth}
    \centering
        \includegraphics[height=2cm]{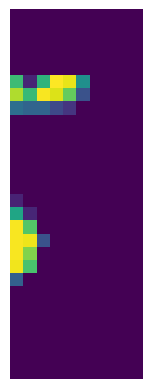}
    \caption{Right Part.}
    \end{subfigure}
    \caption{An example of partitioning a MNIST image and implanting a backdoor trigger.}
    \label{trigger}
\end{figure}
\paragraph{Experimental Setup.} Our evaluation is conducted on five popular image benchmarks: MNIST~\cite{lecun1998gradient}, Fashion\mbox{-}MNIST~\cite{xiao2017fashion}, SVHN~\cite{netzer2011reading}, CIFAR\mbox{-}10 and CIFAR\mbox{-}100~\cite{krizhevsky2009learning}.  
To emulate a VFL scenario, each image is vertically partitioned into three equal slices (left, centre, and right) as set in papers \cite{han2025vertical,wang2024efficient}.  
The centre slice is assigned to the target party to be forgotten. Following the SplitNN paradigm, each client~$P_k$ employs a local encoder~$ML_k$ consisting of two $3{\times}3$ convolutional layers (32 and 64~channels, respectively), each followed by a $2{\times}2$ max-pooling layer. $MG$ first concatenates all client features, then applies two fully connected layers (ReLU hidden size~128) to generate multi-class prediction logits.  
Unless otherwise stated, the hyperparameters are fixed to $c=1.0$ and $\alpha=10^{-3}$. The experiments are conducted on a machine equipped with an Intel(R) Xeon(R) Gold 6240C CPU, 32GB RAM, and an NVIDIA A100-40GB GPUs.

\setlength{\textfloatsep}{\oldtextfloatsep}
\begin{table*}[t!]
\centering
\small
\renewcommand\arraystretch{1}       
\begin{tabular}{lcc|cc|cc|cc|cc}
\bottomrule
\multirow{2}{*}{Method} &
\multicolumn{2}{c|}{MNIST} &
\multicolumn{2}{c|}{Fashion-MNIST} &
\multicolumn{2}{c|}{SVHN} &
\multicolumn{2}{c|}{CIFAR-10} &
\multicolumn{2}{c}{CIFAR-100} \\
& Clean↑ & Bkd.↓ & Clean↑ & Bkd.↓ & Clean↑ & Bkd.↓ & Clean↑ & Bkd.↓ & Clean↑ & Bkd.↓ \\
\hline
Original (Bkd.) & 98.69 & 99.99 & 91.14 & 99.71 & 85.56 & 98.08 & 68.12 & 93.26 & 36.47 & 94.35 \\
Original (no Bkd.) & 98.89 & 9.06 & 91.02 & 9.91 & 86.21 & 9.86 & 68.52 & 9.63 & 36.56 & 1.91 \\
\hline
FedRetrain (Gold) & 91.51 & 9.81 & 88.30 & 10.37 & 70.85 & 9.64 & 60.34 & 10.76 & 30.39 & 1.66 \\
\hline
FedR2S & 89.63 & 18.47 & 86.70 & 18.40 & 64.89 & 16.64 & 54.93 & 18.07 & 19.76 & 27.49 \\
FedGA & 88.42 & 11.09 & 87.02 & 11.10 & 60.74 & 7.87 & 59.15 & 14.01 & \best{26.93} & 22.12\\
FedKD & 87.67 & 14.35 & 86.19 & 12.11 & 61.15 & 8.69 & \best{59.61} & 15.17 & 25.68 & 26.79\\
\textsc{ReMisVFU} (Ours) & \best{90.22} & \best{9.96} & \best{87.08} & \best{10.39} & \best{67.56} & \best{10.22} & 58.45 & \best{10.55} & 25.72 & \best{18.33} \\
\toprule
\end{tabular}
\caption{Clean accuracy (\%) and back-door attack success rate (\%) on five image benchmarks (closer to FedRetrain is better).}
\label{tab:utility_backdoor}
\end{table*}
\begin{table*}[t!]
\centering
\small
\renewcommand\arraystretch{1}       
\resizebox{0.88\linewidth}{!}{%
\begin{tabular}{lcc|cc|cc|cc|cc}
\bottomrule
\multirow{2}{*}{Method} &
\multicolumn{2}{c|}{MNIST} &
\multicolumn{2}{c|}{Fashion-MNIST} &
\multicolumn{2}{c|}{SVHN} &
\multicolumn{2}{c|}{CIFAR-10} &
\multicolumn{2}{c}{CIFAR-100} \\
& AUC. & ACC. & AUC. & ACC. & AUC. & ACC. & AUC. & ACC. & AUC. & ACC. \\
\hline
FedRetrain (Gold) & 0.769 & 0.694 & 0.616 & 0.582 & 0.655 & 0.620 & 0.583 & 0.553 & 0.618 & 0.592 \\
\hline
FedR2S & 0.886 & 0.857 & 0.719 & 0.668 & 0.691 & 0.606 & 0.648 & 0.579 & 0.622 & 0.600\\
FedGA & 1.000 & 1.000 & 1.000 & 1.000 & 1.000 & 1.000 & 1.000 & 1.000 & 1.000 & 1.000 \\
FedKD & 0.887 & 0.849 & 0.742 & 0.658 & 0.755 & 0.655 & 0.750 & 0.666 & 0.721 & 0.657 \\
\textsc{ReMisVFU} (Ours) & \best{0.854} & \best{0.832} & \best{0.697} & \best{0.647} & \best{0.684} & \best{0.599} & \best{0.576} & \best{0.525} & \best{0.609} & \best{0.553} \\
\toprule
\end{tabular}
}
\caption{Membership inference attack success rate on five image benchmarks (closer to FedRetrain is better).}
\label{tab:mia_res}
\end{table*}
\paragraph{Evaluation Metrics.}
We use 4 evaluation metrics widely used in unlearning:
\textbf{1) Clean Accuracy}. The standard classification accuracy obtained on a clean test set (free of injected backdoors) measures model utility. \textbf{2) Backdoor Accuracy}.
To quantify the residual influence of the backdoor and, therefore, the effectiveness of unlearning, we report the proportion of poisoned samples that are classified as the attacker's target label~\cite{han2025vertical,nguyen2024empirical}. A lower value, which is ideally close to the natural class prior, indicates more successful unlearning. We adopt a $2{\times}2$ white trigger stamped in the lower-right corner and set the target label to ``0'' as shown in Figure~\ref{trigger}.  
About $10\%$ of the target party's images are poisoned. \textbf{3) Membership Inference Attack (MIA).}
We further validate unlearning by measuring the success rate of membership inference attacks~(MIA)~\cite{li2025machine,kong2022forgeability}.  
Given the model's output logits with and without the target party, an attacker trains a binary XGBoost classifier~\cite{chen2016xgboost} to decide whether a specific sample participated in VFU.  
A significant drop in MIA success approaching random guessing after unlearning means effective removal of sensitive information. \textbf{4) Runtime Overhead.} This metric captures the runtime cost introduced by the unlearning procedure itself.

\paragraph{Baselines.}
We compare \textsc{ReMisVFU} with 4 representative baselines: \textbf{1) FedRetrain}: fully retrain the VFL model from scratch after removing the target party, which is the ``gold standard".
\textbf{2) FedR2S~\cite{wang2024efficient}}: a rapid re-training scheme that switches between \textsc{RAdam} and momentum~SGD optimisers based on a pre-defined loss threshold.
\textbf{3) FedGA~\cite{han2025vertical}}: gradient-ascent based unlearning that cancels the target party's contribution via adversarial gradient updates.
\textbf{4) FedKD~\cite{varshney2025unlearning}}: knowledge-distillation approach that transfers retained knowledge from the old model to a new one while omitting the target data.

\subsection{Evaluation Results}
\subsubsection{Utility and Forgetting Performance.}
Table \ref{tab:utility_backdoor} compares \textsc{ReMisVFU} with baselines on the joint goals of utility (clean accuracy) and forgetting effectiveness (back-door success rate).
Across the 5 datasets, the clean-accuracy of \textsc{ReMisVFU} is only 2.45\% lower than the fully retrained ``gold" model, while FedR2S, FedGA, and FedKD are on average 5.10\%, 3.83\%, and 4.22\% lower than the retrained model, respectively. In some scenarios, although ReMisVFU is slightly weaker, its privacy performance is superior. Its attack-success rates are closest to the natural class priors in ``Original (no Bkd.)" rows and are either the best on every dataset. By contrast, FedR2S leaves substantial back-door residue, increasing it by an average of 7.84\%, showing that switching optimizers after loading a pretrained model is insufficient for thorough forgetting. FedGA suppresses triggers to roughly 9–14\% on easier datasets but deteriorates to 22.1\% on CIFAR-100, while FedKD's performance falls between the FedR2S and FedGA. To sum up, \textsc{ReMisVFU} achieves a more complete erasure of the forgotten features and, combined with its competitive utility, demonstrates the most robust overall performance among all baselines.

\subsubsection{Privacy (Membership Inference).}
We measure post‑unlearning privacy leakage with the MIA success rate. The larger the value, the easier it is for an adversary to decide whether a sample participated in training. As is customary, we report AUC (area under the ROC curve) and ACC (attack accuracy); the closer these scores are to FedRetrain, the better the unlearning quality. Table \ref{tab:mia_res} shows that \textsc{ReMisVFU} attains MIA metrics closest to FedRetrain on every dataset: on average, the AUC gap between the two methods is only 4.22\% (and 5.82\% for ACC). In comparison, FedKD and FedR2S exhibit average AUC gaps of 12.28\% and 6.50\% (ACC gaps of 8.89\% and 5.94\%, respectively). FedGA, however, performs catastrophically on all MIA privacy indicators because its gradient‑ascent ``forgetting" step drives the model into a highly over‑fitted, sharp region of the loss landscape.

\begin{figure}[t]
  \centering
  \setlength{\tabcolsep}{2pt} 
  \renewcommand{\arraystretch}{1.0}
  \begin{tabular}{cc}
    \subcaptionbox{MNIST – Original\label{fig:mnist-orig}}[0.48\linewidth]{
      \includegraphics[width=\linewidth]{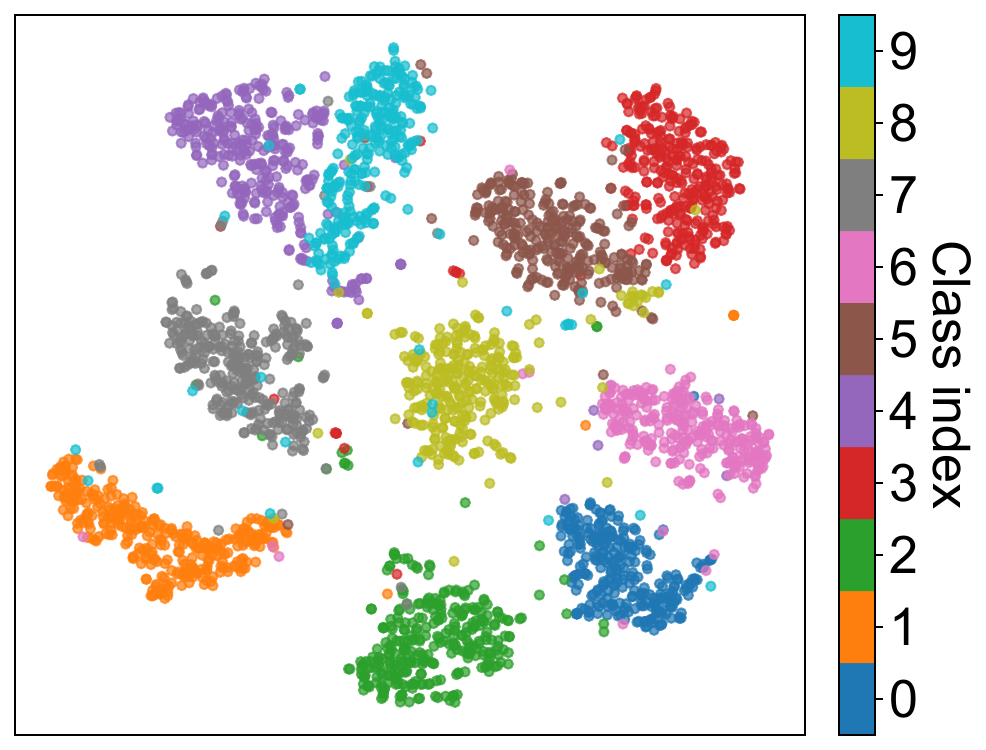}} &
    \subcaptionbox{FashionMNIST – Original\label{fig:fashion-orig}}[0.48\linewidth]{
      \includegraphics[width=\linewidth]{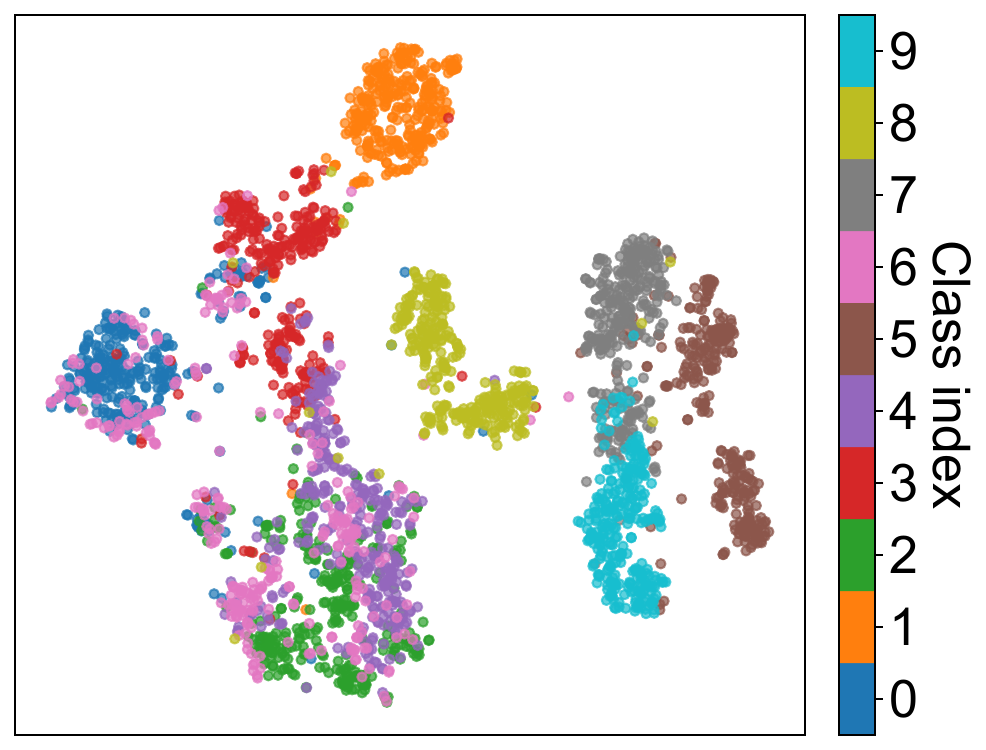}} \\
    
    \subcaptionbox{MNIST – Retrained\label{fig:mnist-retrain}}[0.48\linewidth]{
      \includegraphics[width=\linewidth]{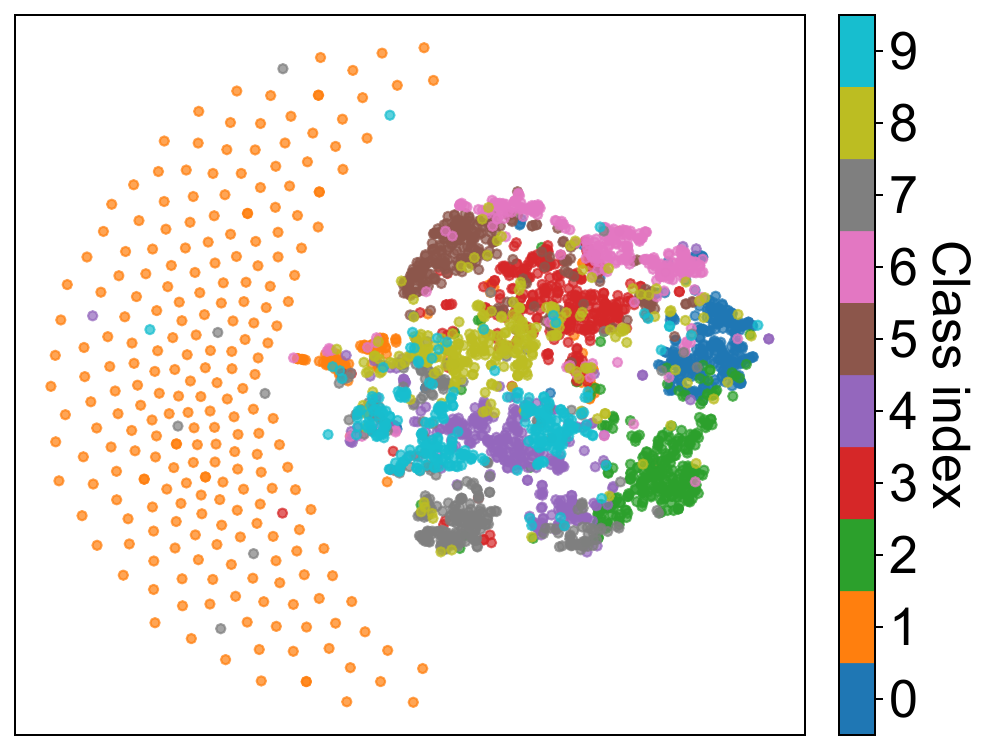}} &
    \subcaptionbox{FashionMNIST – Retrained\label{fig:fashion-retrain}}[0.48\linewidth]{
      \includegraphics[width=\linewidth]{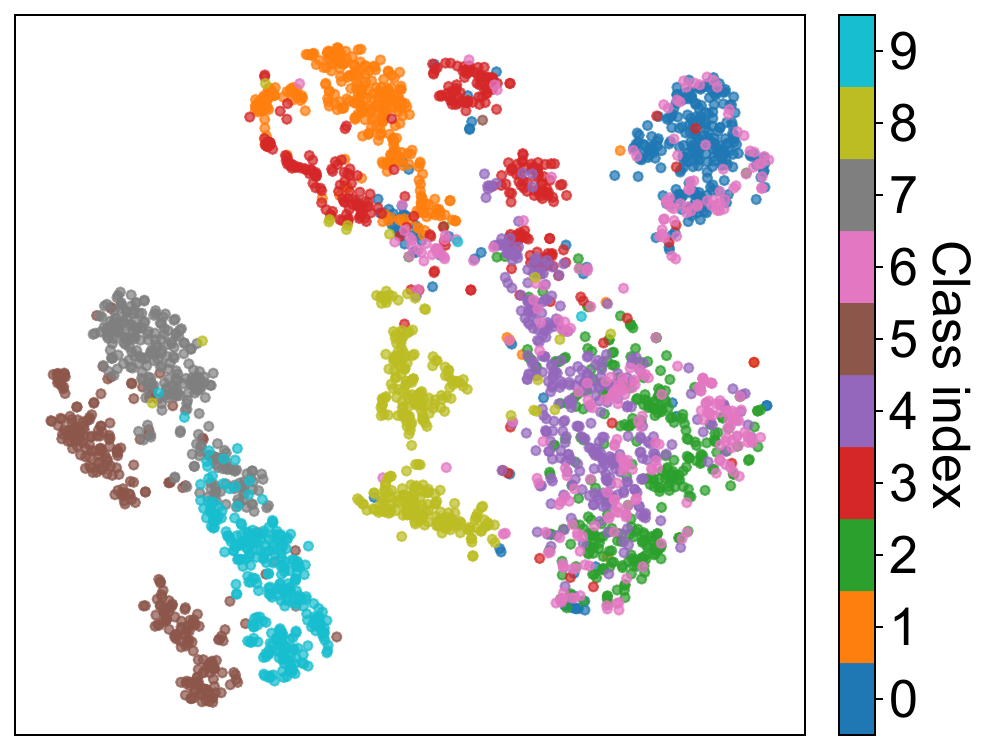}} \\

    \subcaptionbox{MNIST – Unlearned\label{fig:mnist-unlearn}}[0.48\linewidth]{
      \includegraphics[width=\linewidth]{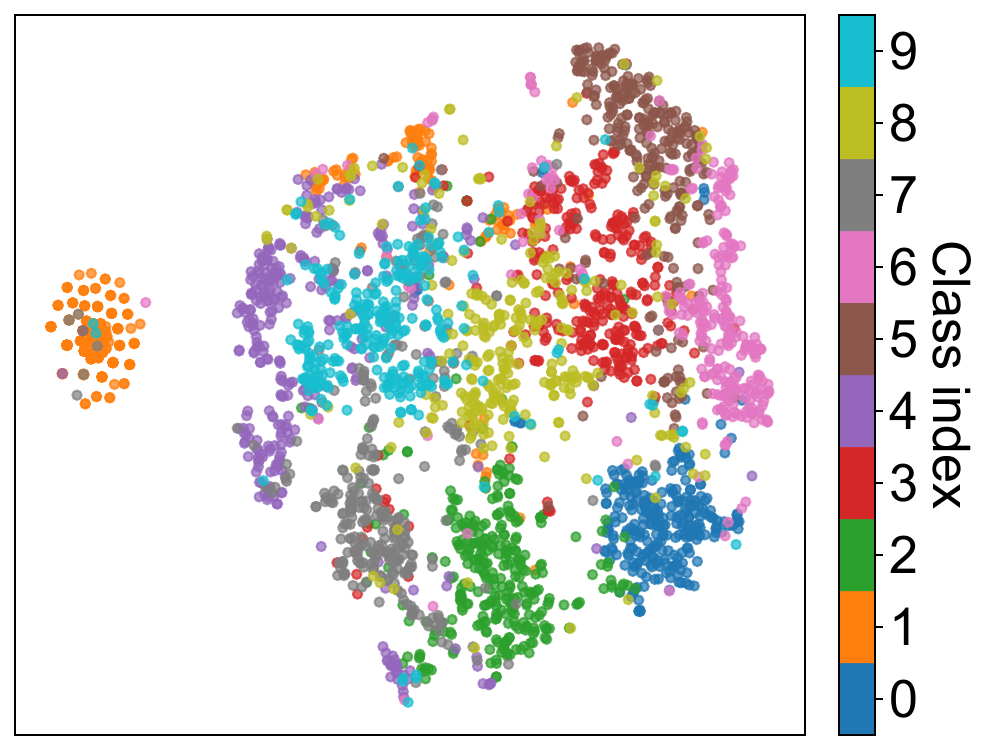}} &
    \subcaptionbox{FashionMNIST – Unlearned\label{fig:fashion-unlearn}}[0.48\linewidth]{
      \includegraphics[width=\linewidth]{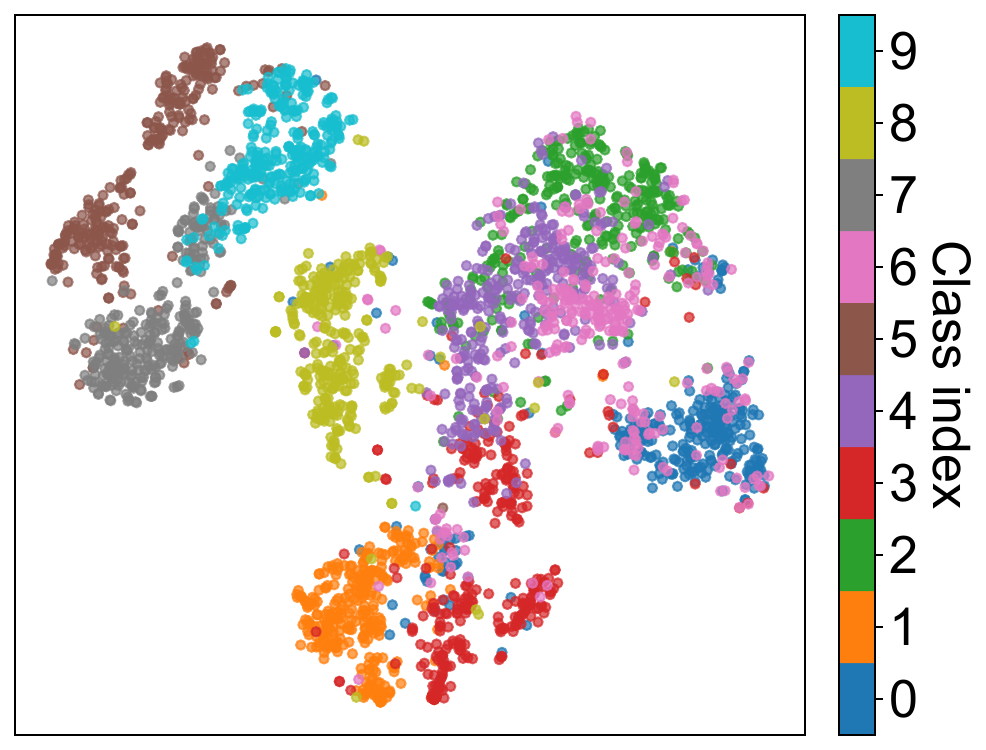}} \\
  \end{tabular}

  \caption{t‑SNE visualization of the intermediate features $\{h_{i,k}\}_{k=1}^{K-1}$, with different colors denoting different classes.}
  \label{fig:tsne_all}
\end{figure}

\subsubsection{Visualization.}
To illustrate the effect of \textsc{ReMisVFU} from a representational viewpoint, we ran t‑SNE visualization \cite{maaten2008visualizing} on the intermediate VFL features $\{h_{i,k}\}_{k=1}^{K-1}$ for MNIST and Fashion‑MNIST. As shown in Figure~\ref{fig:tsne_all}, after only a few iterations, the resulting feature distribution closely matches that of the fully retrained Gold Baseline, but with a more uniform density. This shows that our representation‑misdirection strategy can selectively erase the semantic information contributed by the forgotten party while preserving the discriminative power of the remaining parties, which removes the forgotten party's influence from deeper layers of the model.

\subsubsection{Efficiency (Running Overhead).}
\begin{figure}[t!]
    \centering
    \includegraphics[width=1\linewidth]{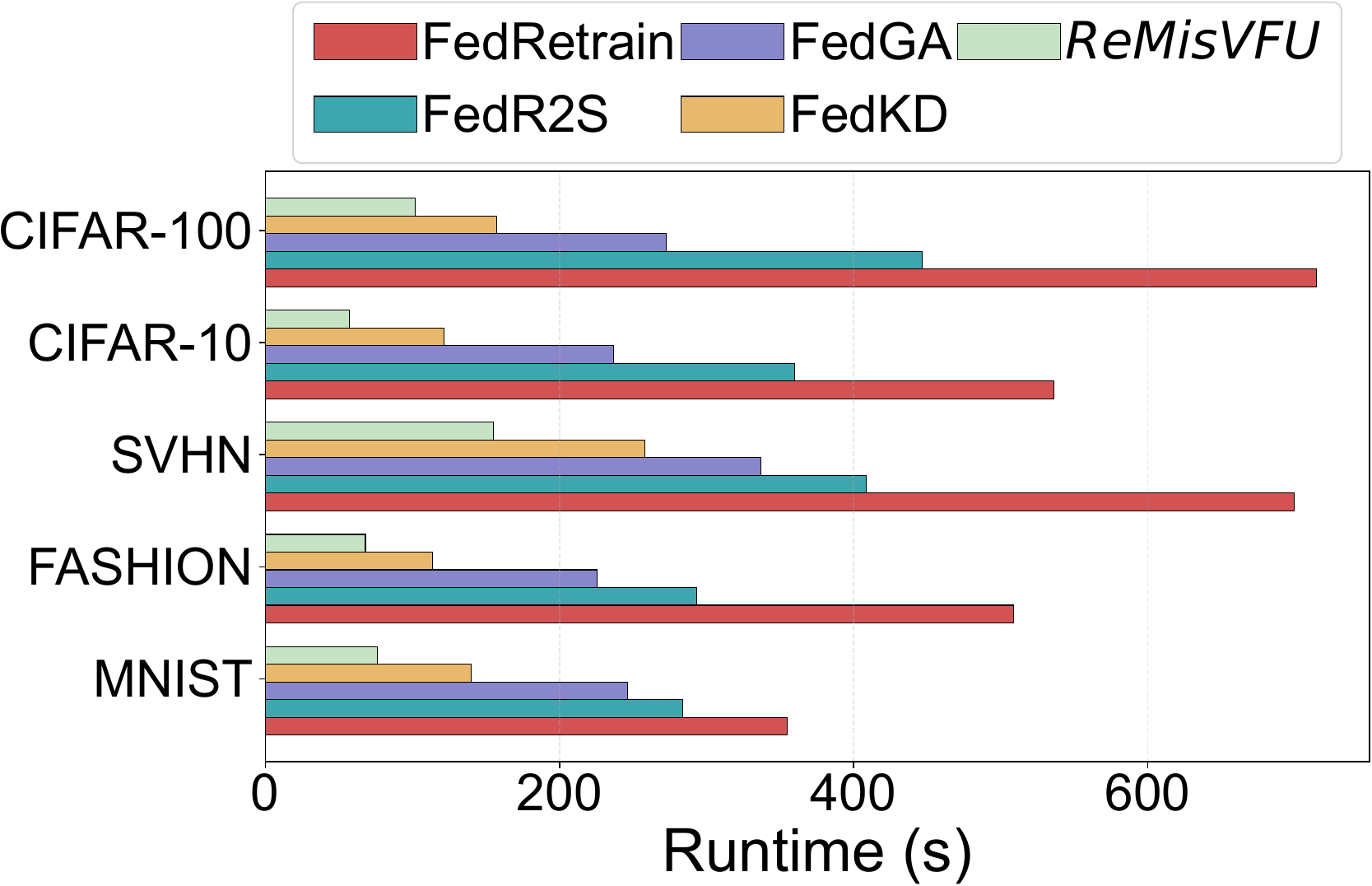}
    \caption{Comparison of the VFU training time.}
    \label{fig:runtime}
\end{figure}

To highlight the runtime efficiency, Figure \ref{fig:runtime} presents end-to-end unlearning times on each benchmark. \textsc{ReMisVFU} completes forgetting using only 10.63\%–21.49\% of FedRetrain's time, representing a substantial speed-up. Compared with FedR2S, despite both methods avoiding full retraining, \textsc{ReMisVFU} still cuts wall-clock time by 24.33\%, chiefly because it removes the costly optimizer-switching heuristic and attains faster convergence through multi-round feature masking. FedGA also converges more slowly, requiring roughly $3.89\times$ our runtime, as its adversarial ascent steps frequently push the model into high-loss regions, which must later be corrected with additional fine-tuning rounds. FedKD employs a two-stage teacher–student paradigm; the mandatory forward pass of the teacher network in every round incurs extra overhead, resulting a runtime about $1.77\times$ that of \textsc{ReMisVFU}. In summary, by editing internal representations instead of retraining entire networks, \textsc{ReMisVFU} achieves markedly lower time cost.

\subsubsection{Ablation Study.}
To quantify the contribution of our gradient--projection strategy in Eq.~(\ref{cord})--(\ref{gradall}), we compare \textsc{ReMisVFU} with two degraded variants: i) \emph{No‑GCM}, which jointly optimises \(\mathcal L_f\ +\ \alpha \mathcal L_r\) using vanilla gradient summation in Eq.~(\ref{origrad}); ii) \emph{Rand‑Proj}, which replaces the orthogonal projection with a random unit vector of the same norm, serving as a sanity check that the benefit indeed stems from conflict resolution rather than mere gradient scaling. Table~\ref{tab:gradablation} reports that \emph{No‑GCM} raises the back‑door attack success rate by 2.47\% and lowers the clean accuracy by 1.71\% on average, demonstrating that naively summing gradients introduces destructive interference. Random projection further causes a pronounced degradation in both forgotten effectiveness and retained utility. With GCM enabled, our method consistently attains better utility–privacy trade‑off, confirming that explicit gradient alignment is essential for VFU.
\subsubsection{Parameter Sensitivity Analysis.}
\begin{figure}[t!]
    \centering
    \begin{subfigure}{.48\linewidth}
        \centering
        \includegraphics[width=\linewidth]{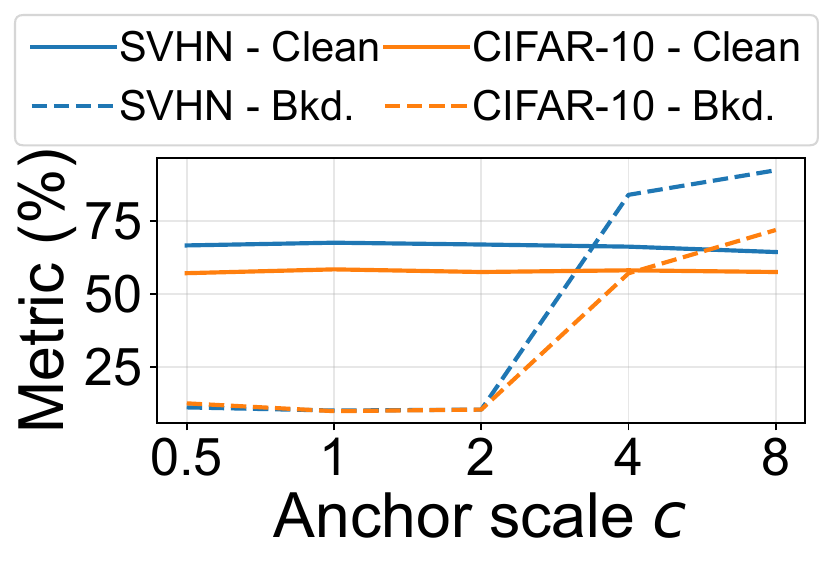}
        \caption{Varying \(c\)}
        \label{fig:sensitivity_c}
    \end{subfigure}
    \hfill
    \begin{subfigure}{.48\linewidth}
        \centering
    \includegraphics[width=\linewidth]{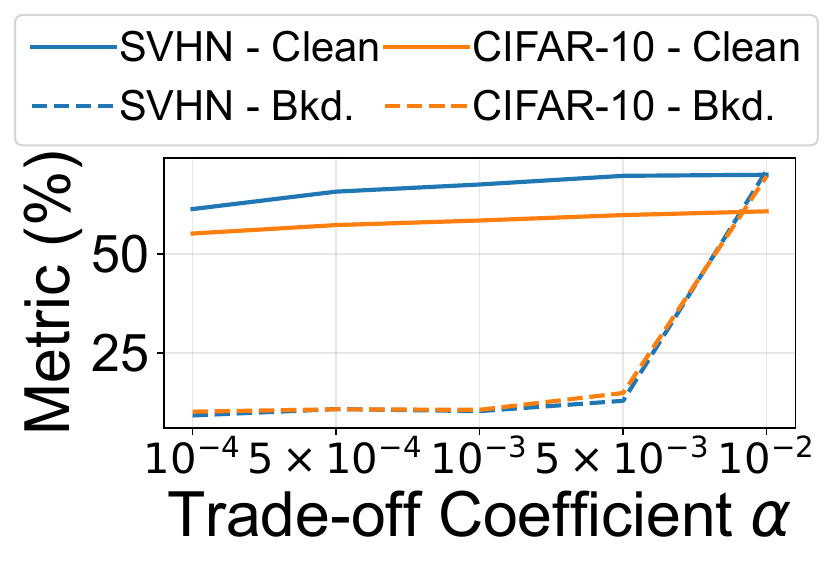}
        \caption{Varying \(\alpha\)}
        \label{fig:sensitivity_alpha}
    \end{subfigure}
    \caption{Parameter sensitivity on SVHN and CIFAR‑10.}
    \label{fig:sensitivity}
\end{figure}

We investigate the influence of anchor scaling factor \(c\) in Eq.~(\ref{loss1}) and the trade‑off coefficient \(\alpha\) in Eq.~(\ref{gradall}), respectively. Figure~\ref{fig:sensitivity_c} sweeps \(c\in\{0.5,1,2,4,8\}\) while fixing \(\alpha\!=\!10^{-3}\); Figure~\ref{fig:sensitivity_alpha} varies \(\alpha\in\{10^{-4},5\!\times\!10^{-4},10^{-3},5\!\times\!10^{-3},10^{-2}\}\) with \(c\!=\!1\). For the anchor scale \(c\), when $c \le 2$, both clean accuracy and back‑door success remain essentially flat, indicating that moderate feature compression does not hinder model utility. Once $c \ge 4$, the degree of forgetting drops sharply, because an excessively large anchor pushes intermediate features into the nonlinear saturation regime of the activation functions, diluting the forgetting signal and allowing the trigger to re‑emerge. For the trade‑off coefficient \(\alpha\), when $\alpha$ lies in the range $1\times10^{-4}\le\alpha\le1\times10^{-3}$, clean accuracy rises slightly with increasing $\alpha$, while the back‑door success rate remains near the random‑guessing baseline. Once $\alpha$ surpasses $5\times10^{-3}$, the back‑door success on both datasets quickly exceeds 50\%, indicating that the retention term overwhelms the forgetting objective. Balancing utility and privacy, we therefore adopt $\alpha = 10^{-3}$ as the default setting.
\begin{table}[t!]
\centering
\setlength{\tabcolsep}{4pt}
\resizebox{\linewidth}{!}{%
\begin{tabular}{lcccc}
\toprule
\multirow{2}{*}{Method} &
\multicolumn{2}{c}{SVHN} &
\multicolumn{2}{c}{CIFAR‑10} \\
\cmidrule(lr){2-3}\cmidrule(lr){4-5}
 & Clean Acc.\(\uparrow\) & Back‑door\(\downarrow\) &
   Clean Acc.\(\uparrow\) & Back‑door\(\downarrow\) \\
\midrule
No‑GCM       & 65.04{\small\(\pm0.85\)} & 11.65{\small\(\pm1.45\)}  & 57.54 {\small\(\pm1.24\)} & 14.07 {\small\(\pm0.41\)} \\
Rand‑Proj    & 35.62 {\small\(\pm2.86\)} & 82.63 {\small\(\pm3.45\)}  & 32.07 {\small\(\pm0.19\)} & 87.21 {\small\(\pm4.38\)} \\
\textsc{ReMisVFU} & \textbf{67.56} {\small\(\pm1.02\)} & \textbf{10.22} {\small\(\pm0.56\)} & \textbf{58.45 {\small\(\pm1.17\)}} & \textbf{10.55 {\small\(\pm0.23\)}} \\
\bottomrule
\end{tabular}
}
\caption{Effect of gradient conflict mitigation.}
\label{tab:gradablation}
\end{table}

\section{Conclusion}
In this paper, we present \textsc{ReMisVFU}, a plug‑and‑play VFU framework that cuts off a forgotten party's influence by misdirecting intermediate representations and reconciles the trade‑off between forgetting and retention via orthogonal gradient projection. Extensive evaluations show that through this combination, it achieves an efficient balance between privacy‑preserving unlearning and strong model utility.

\section{Acknowledgments}
This work was supported by the National Natural Science Foundation of China (62341410, 62302348), National Key Research and Development Program of China (2023YFE0205700) and the State Key Laboratory of Internet of Things for Smart City (University of Macau) Open Research Project (SKL-IoTSC(UM)/ORP05/2025).
\bibliography{aaai2026}
\end{document}